\documentclass[letterpaper, 10 pt, conference]{ieeeconf}  %

\IEEEoverridecommandlockouts                              %

\usepackage[T1]{fontenc}
\usepackage[backend=biber, sorting=none]{biblatex}
\usepackage{graphicx} 
\usepackage{amsmath}
\usepackage{authblk}
\usepackage{algorithm}
\usepackage{algpseudocode}
\usepackage{caption}
\usepackage{subfig}
\usepackage{amssymb}
\newcommand\doubleplus{+\kern-1.3ex+\kern0.8ex}
\algnewcommand\algorithmicforeach{\textbf{for each:}}
\algnewcommand\ForEach{\item[ \algorithmicforeach]}
\renewcommand{\vec}[1]{\mathbf{#1}}
\usepackage[font=small,labelfont=]{caption}

\title{\LARGE \bf
Spatio-Semantic ConvNet-Based Visual Place Recognition
}

\author{Luis G. Camara and Libor P\v{r}eu\v{c}il
\thanks{\hspace*{-1em}\bf{978-1-7281-3605-9/19/\$31.00 \textcopyright \,2019 IEEE}}%
}  %
\affil{\small Czech Institute of Informatics, Robotics and Cybernetics, Czech Technical University in Prague, Prague, Czech Republic}
\affil{\tt \small luis.gomez.camara@cvut.cz}

\bibliography{bibliography}
\begin{document}
\newcommand{\subf}[2]{%
{\small\begin{tabular}[t]{@{}c@{}}
#1\\#2
\end{tabular}}
}

\graphicspath{{Figures/}}

\maketitle
\thispagestyle{empty}
\pagestyle{empty}

\begin{abstract}

We present a Visual Place Recognition system  that follows the two-stage format common to image retrieval pipelines.  The system  encodes images of places by employing the activations of different layers of a pre-trained, off-the-shelf, VGG16 Convolutional Neural Network (CNN) architecture. In the first stage of our method and given a query image of a place,  a number of top candidate images is retrieved from a previously stored  database of places. In the second stage, we propose an exhaustive comparison  of the query image against these candidates by encoding semantic and spatial information in the form of CNN features. Results from our approach outperform by a large margin state-of-the-art visual place recognition methods on five of the most commonly used benchmark datasets. The performance gain is especially remarkable on the most challenging datasets, with more than a twofold recognition improvement with respect to the latest published work. 

\end{abstract}

\section{INTRODUCTION}

\begin{figure}[htbp]
\framebox{\parbox{2.0in}
\centering
\includegraphics[scale=0.21]{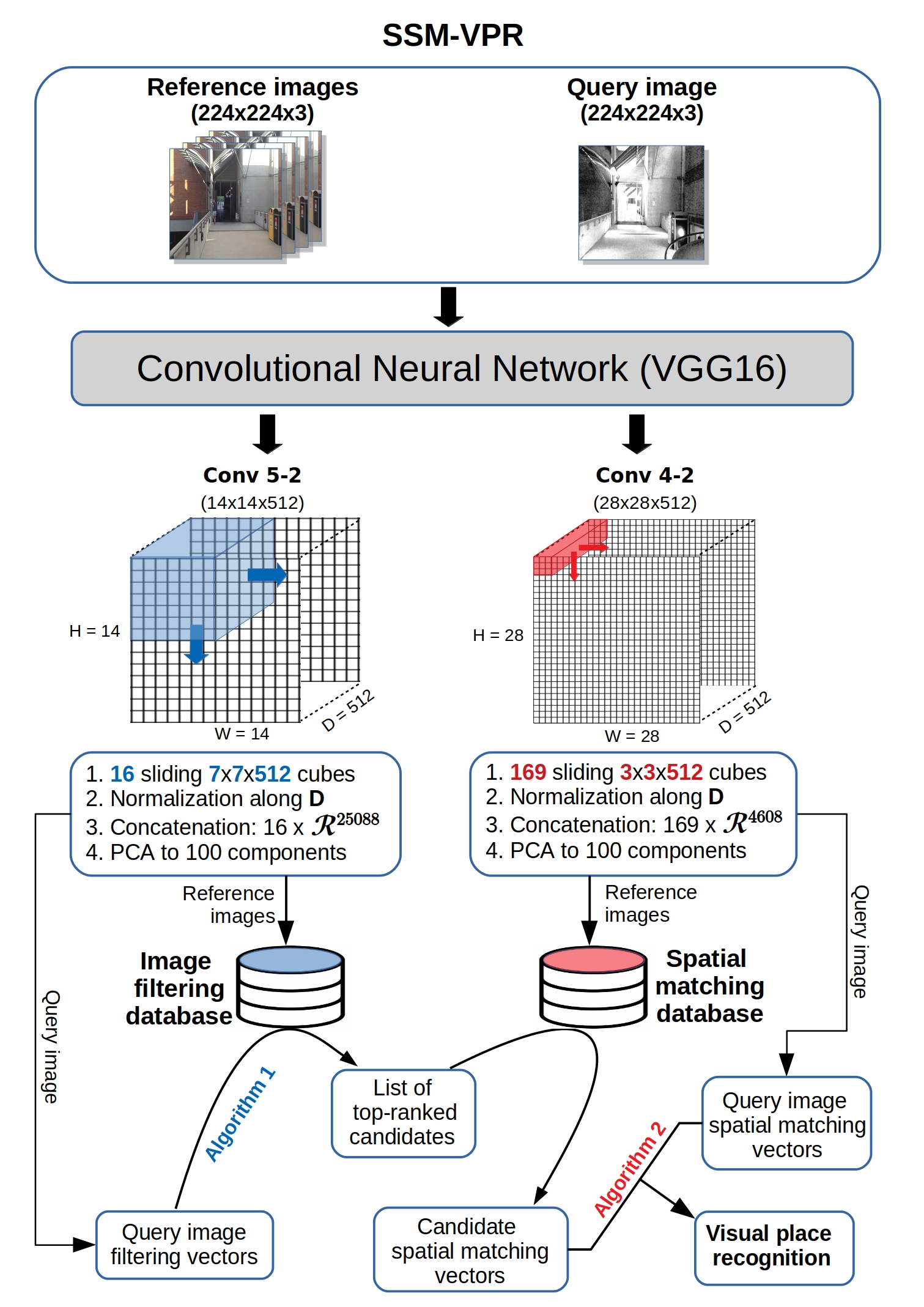}}
\caption{Workflow of our Semantic and Spatial Matching Visual Place Recognition (SSM-VPR) system.} 
\vspace*{-6mm}
\label{fig_workflow}
\vspace*{-0mm}
\end{figure}  
In the field of mobile robotics and particularly in applications requiring long-term navigation and localization\,\cite{tipaldi2013lifelong, churchill2012practice%
}, the problem of \textit{Visual Place Recognition} (VPR) is a fundamental one and still an open computer vision  research topic\,\cite{lowry2016visual, galvez2012bags, arandjelovic2016netvlad, torii2013visual}. Navigation systems used by mobile robots are typically based on \textit{Simultaneous Localization And Mapping} (SLAM) approaches\,\cite{durrant2006simultaneous, montemerlo2002fastslam, davison2003real}. In long-term SLAM implementations,  the task of determining whether a robot is located at a previously visited place becomes a basic requirement. This is known as \textit{loop closure} and   is typically solved by VPR\,\cite{galvez2012bags, hou2015convolutional, angeli2008fast%
} in image-based systems. 

Visual place recognition is however a challenging problem, especially in uncontrolled outdoors environments and over long periods of time. Images captured by a robot at a specific location can differ significantly with respect to those captured on a first pass through the same location. This is due to environmental factors such as differences in the season of the year, changing weather conditions, day-night cycles, light variations during the day and/or purely geometric aspects such as changes in viewpoint between two traverses.   

Loop closure is typically approached in the literature by using  Bag of (visual) Words (BoW) models\,\cite{galvez2012bags,angeli2008fast, cummins2008fab}   
or the closely related VLAD\,\cite{arandjelovic2016netvlad, torii201524, lowry2018lightweight} and Fisher vectors\,\cite{perronnin2010large, douze2011combining}.  These approaches aggregate robust image features into previously created dictionaries of visual words,  leveraging  the resulting compact representations for the creation and fast query  of image databases. Nonetheless, these models are strictly content-based and do not take into account spatial relationships between image features. It is for this reason that they are usually fairly robust to viewpoint changes but suffer from problems such as perceptual aliasing\,\cite{kejriwal2016high}.

More recently, inspired by the great success of Convolutional Neural Networks (CNN) in several computer vision tasks, a number of authors have employed the activations of some  CNN layers to create image representations suitable to tackle the VPR problem\,\cite{arandjelovic2016netvlad, hou2015convolutional, chen2014convolutional, sunderhauf2015performance, sunderhauf2015place, arroyo2016fusion, chen2017deep, chen2017only, khaliq2018holistic}. By discarding the fully connected layers of CNNs and using the output of their middle and later convolutional layers instead, it has been shown that it is possible to encode rich semantic information  that  may be robust to several image changes. Yet, most  authors do not explicitly take into consideration the spatial locations of the activations and aggregate them into orderless  BoW models or simply concatenate them into high-dimensional spaces.  

In this work, we take a step forward and approach the VPR problem by focusing not only on the semantics provided by CNN features but also, and very importantly, on the geometric relationships between these features in images. Some recent works\,\cite{rocco2017convolutional, taira2018inloc} have in fact considered CNNs to geometrically match images, although outdoor VPR under challenging conditions has not been explicitly addressed. The main contribution of our work is the implementation and evaluation of a VPR system based on a very simple yet effective CNN-based spatial matching stage, which is preceded by a baseline image retrieval stage, also based on CNN features. At present, we focus on visual place recognition performance rather than optimizing computational complexity and show that our approach  outperforms  by a large margin  previously published results on five benchmark image datasets\,\cite{chen2017only, khaliq2018holistic}.

The remaining of this paper is structured as follows. In Section~\ref{related_work} we review published work related to the VPR problem and put our work into context. Our methodology is presented in Section~\ref{methodology}, characterizing  our ground truth and describing the two stages of our system. In Section~\ref{results}, experimental results are presented, compared with the state-of-the-art and discussed. Finally, Section~\ref{conclusions} is left to conclusions and to describe our line of work in the future.

\section{RELATED WORK}\label{related_work}
Previous to the resurgence of CNN models\,\cite{srinivas2016taxonomy}, commonly followed computer vision approaches in VPR employed handcrafted robust features such as SIFT\,\cite{lowe1999object},  SURF\,\cite{lowe1999object}, ORB\,\cite{rublee2011orb}, etc. to represent images,  encoding them into BoW-like models by using pre-trained dictionaries of visual words\,\cite{galvez2012bags,  angeli2008fast, torii201524, filliat2007visual}. 
In recent years, however, CNN-based features have shown their superiority in many computer vision tasks\,\cite{simonyan2014very, girshick2014rich, badrinarayanan2017segnet}, including VPR. Hence, latest published work in VPR is mostly based on CNN image representations and also our main focus in this section.

Some of the first works utilizing CNN descriptors in the VPR problem can be found in  \cite{chen2014convolutional, sunderhauf2015performance}, where the authors used pre-trained networks and explored the capabilities of different layers on different aspects of the place recognition task. They observed that later layers held more semantic information and therefore were more robust under conditions of large viewpoint variance, whereas middle layers were less affected by appeareance changes such as illumination.  Hou et al.\,\cite{hou2015convolutional} also investigated the performance of CNN  intermediate layers as image descriptors and concluded that they were faster to compute and achieved similar or better performance compared to handcrafted image descriptors.  In\,\cite{sunderhauf2015place}, the authors extracted potential landmarks from images using the \textit{Edge Boxes}\,\cite{zitnick2014edge} object proposal algorithm and tried to find the best matches between them. They significantly improved recognition performance,  although  the detector itself imposed a heavy computational load. 
Arroyo et al.\,\cite{arroyo2016fusion} proposed a methodology where  the activations of multiple convolutional layers were fused by concatenation, compression and binarization, showing better performance than traditional image descriptors and even some baseline CNN approaches.

Instead of using pre-trained architectures, some authors have trained their networks specifically for the VPR task. For instance, \cite{chen2017deep} created a  large dataset of places  and trained a network that interpreted VPR as a classification problem, achieving an average 10\,\% increase in performance over other approaches. Of great  relevance  in VPR is the work of Arandjelovi\'c et al.\,\cite{arandjelovic2016netvlad}, who designed a CNN architecture, NetVLAD,  which incorporated a VLAD layer and could be trained in an end-to-end fashion for the place recognition task. Their results on some very challenging datasets were remarkable, significantly outperforming state-of-the-art works based on pre-trained CNNs.

Recently, following the success of region-based approaches, Chen et al.\,\cite{chen2017only} created a system that used a late convolutional layer as a landmark detector and an earlier one to create local descriptors to match the detected landmarks. Their system showed improved recognition under strong viewpoint and condition variations. One of the latest published work in VPR is that of Kaliq et al.\,\cite{khaliq2018holistic}. By detecting regional features and VLAD-encoding them, they created a lightweight pipeline that outperformed the 2-layer approach of \cite{chen2017only}. 

More in  line with our two-stage methodology, some authors have in fact considered geometric post-verification of shortlisted locations\,\cite{cummins2011appearance,cadena2012robust},
 resulting in a significant boost in recognition performance. These authors focused however on handcrafted features.   
To the best of our knowledge, a system that takes advantage of the power of CNN features for both retrieval and geometric verification in the context of VPR is lacking in the literature. 

\section{PROPOSED METHODOLOGY}\label{methodology}

We consider the problem of VPR as an \textit{image retrieval} one\,\cite{arandjelovic2016netvlad, sattler2016large, mohedano2016bags}, which is  usually divided into (i) a \textit{filtering stage} and (ii) a \textit{re-ranking} stage\,\cite{tolias2015particular}. During stage (i) and given a \textit{query image}, a whole database of \textit{reference images} is searched and a  number of top-ranked candidates retrieved according to some similarity or distance metric. In stage (ii), the candidates are compared with the query image in a more exhaustive fashion and re-ranked to give an ordered list of the bests matches. The place recognition problem then reduces to choosing the top match from the re-ranked list.

In our approach, we follow this two-stage structure and encode our images using two different layers of the VGG16 CNN architecture\,\cite{simonyan2014very}, one for each stage. We employed the VGG16 network pre-trained on the Places205 dataset\,\cite{zhou2014learning}, which is conceptually  closer to VPR than the more commonly used ImageNet dataset\,\cite{deng2009imagenet}, oriented towards object recognition (we advance that it is our plan to explore in the future end-to-end trainable networks such as NetVLAD).
For convenience, we called our approach SSM-VPR (Semantic and Spatial Matching Visual Place Recognition).

The structure of our pipeline is illustrated in Figure\,\ref{fig_workflow}. Reference and query images are passed through the network and the activations of two specific layers, \textit{conv\,5-2} and \textit{conv\,4-2}, stored for processing. In the case of reference images, permanent databases are created for later comparison against query images. Our choice of layers is based on experimentation. Layer \textit{conv\,5-2} contains less spatial resolution (14\,$\times$\,14 activations per feature map) although, being one of the latest layers in the network, provides strong semantics. This materializes into better performance when used in the image filtering stage. On the other hand, layer \textit{conv\,4-2} contains twice the spatial resolution. It is therefore more suitable for geometrical comparisons between images  while still producing features that are robust enough to image changes. Each layer contains $D\!\!=\!\!512$ feature maps and a total of H$\times$W$\times$D activations, where the first two dimensions correspond to spatial locations. All images were scaled to size $224\!\times\!224\!\times\!3$ RGB.

Before describing the two stages of our system in more detail, we first comment on the way our ground truth is defined, since this directly affects the presented recognition results.

\subsection{Ground truth and distance tolerance}\label{ground_truth} 
During the evaluation of a VPR system, it must be decided whether the guessed location of a query image is a true positive or not. Since the definition of a place is somehow fuzzy, a distance tolerance is highly convenient for this task. We have adopted from\,\cite{arandjelovic2016netvlad, torii201524, arandjelovic2014dislocation} 
 the criterion that a query image is deemed as a true positive when its ground truth location is within 25\,m from the guessed location.

The datasets employed throughout this work consist of pairs of image sequences of the same traverse but under different conditions (see Section\,\ref{datasets}). They have been tagged so that frames corresponding to the same place in both traverses contain the same numerical label. This allows for the use of frame tolerance instead of distance in the evaluation of true positives.  
We propose to find the optimal frame tolerance in the datasets by directly looking at the recognition performance of the system as the tolerance is varied. 
Hence, an analysis of performance vs. frame tolerance is carried out in Section~\ref{results}.

\subsection{Image filtering stage}\label{image_retrieval_method}
We want to make our image filtering stage as robust to viewpoint changes as possible. Using full-image representations can rapidly degrade recognition performance under changes in viewpoint, as large parts of the reference images may be missing in the query image. Several authors have shown that selecting regions of interest can boost recognition performance\,\cite{sunderhauf2015place, chen2017only, khaliq2018holistic}. Rather than trying to find those regions, we take a brute force approach and spatially scan the feature maps  by dividing   them into smaller sections. 

Based on layer \textit{conv\,5-2} and as seen in Figure~\ref{fig_workflow}, cubes of dimensions  $7\!\times7\!\times\!512$  were defined by sliding along the layer's horizontal (W) and vertical (H) directions. With a stride size of $s\!=\!2$ activations, the sliding process led to  a total of 16 cubes per image.  
After normalization along the direction of the feature maps, we concatenated the activations in each cube and created 16 vectors of dimension $7\!\times\!7\!\times\!512\!=\!25088$. Since working in such a high dimensional space is expensive in terms of memory and complexity, we performed dimensionality reduction through Principal Component Analysis (PCA), a common practice when working with CNN features\,\cite{chatfield2014return}.  

As it will be shown in  Section~\ref{datasets} and in order to justify our use of PCA, we conducted several experiments on the considered datasets and observed that the recall@25  (the percentage of successfully retrieved images when using a list of candidates of size 25) did not improve above 80 principal components. The criteria of success was that at least one image in the list was close enough to the ground truth, with a tolerance of $\pm2$ frames. Prudently, we decided to reduce  our vectors to $d\!=\!100$ dimensions instead of 80. For each image in the reference dataset, we encoded the convolutional cubes and stored them in an \textit{image filtering database} (IFDB). We did not store any information about the spatial location of the cubes. 

The same processing described above was performed on any entering query image. Subsequently, the vectors representing that image were compared one by one against the IFDB and, based on a nearest neighbor distance criterion, the first $N$ candidates were added to a histogram of places. The resulting accumulated  histogram of places was then used to extract a list of $N$ top-ranked candidate images (see Algorithm \ref{alg1} for a pseudo code).      
\begin{algorithm}
\caption{Image filtering stage}
\begin{algorithmic}[1]
\small
\State Get query image vectors $\{\vec{q}_{i}\}$ (16$\times \mathbb{R}^{100}$) from $conv5\_2$ 
\State $cand\_hist \gets$Initialize histogram of candidate images
\For {$i=1$ to 16}
	\State Find best $N$ matches of $\vec{q}_{i}$ in IFDB
	\State Update the corresponding $N$ bins in $cand\_hist$ 
\EndFor
\State Candidate list $\gets$Select $N$ highest bins from $cand\_hist$
\end{algorithmic}
\label{alg1}
\end{algorithm}
\vspace*{-4mm}
\subsection{Spatial matching stage}
The outcome of the filtering stage is an ordered list of $N$ images that are similar to the query image in a vector distance sense. We want to emphasize the notion that the geometric relationships or spatial constrains among the elements of images are of vital importance in the recognition of a place. We believe humans recognize places by identifying semantic elements in a scene and then finding geometric relationships between them\,\cite{lynch1960image}. Following this intuition, we created a matching system based on convolutional features that is able to capture  semantic information and at the same time  keep track of the spatial relationships between those features. 
\begin{figure}
\framebox{\parbox{2.0in}
\centering
\includegraphics[scale=0.21]{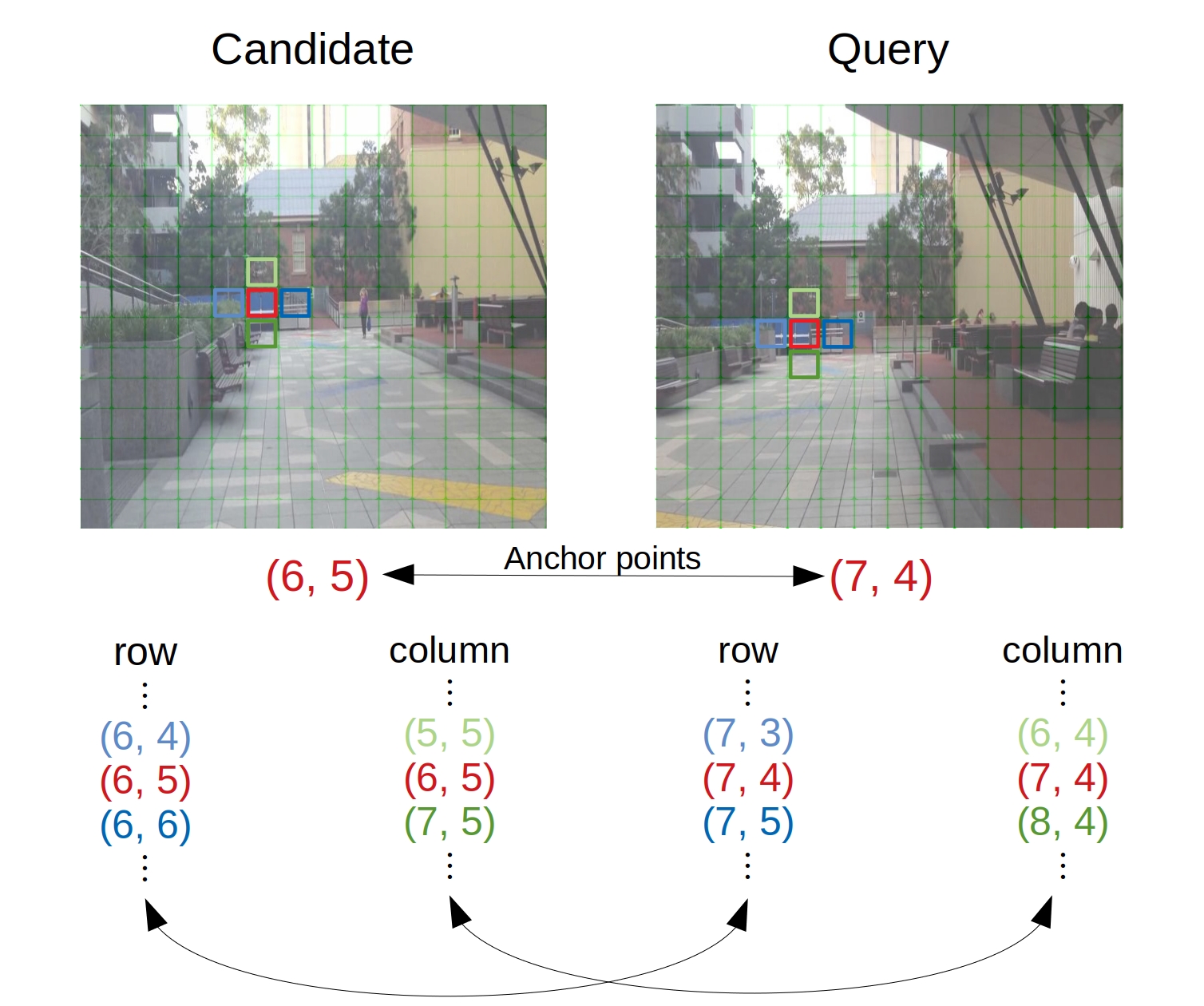}}
\caption{Illustration of the spatial matching approach for an individual location (square-marked region) within a candidate image.} 
\label{fig_spatial_matching}
\vspace{-4mm}
\end{figure} 

As can be appreciated from Figure\,\ref{fig_workflow}, the number of activations in each convolutional cube for layer \textit{conv\,4-2} is $3\!\times\!3\!\times\!512\!=\!4608$. After normalization, concatenation, PCA compression and the application of a stride of $s\!=\!2$,  each image is represented  by an array of $13\!\times\!13\!=\!169$ vectors in $\mathbb{R}^{100}$.  For each image, we stored those vectors in a \textit{spatial matching database} (SMDB),  along with their position in the array.  In this fashion, when comparing query and candidate images, it is possible to match both  convolutional features and their spatial arrangement. 
In what follows, we  provide the main ideas of our spatial matching approach, which are also summarized as pseudo code in Algorithm \ref{alg2}.

We take the list of $N$ candidates provided by the first stage of our system and compare  each one of them with the query image. 
Given a candidate image,  we retrieve from the SMDB the collection of vectors $\{\vec{c}_{i,j}\}$ representing it as well as their locations $(i,j)$. A spatial consistency check against the query image representation, $\{\vec{q}_{k,l}\}$,  is then performed. This is schematically illustrated in Figure\,\ref{fig_spatial_matching}, where the grids seen in the two images define the spatial locations of the vectors. Let us focus for instance on position $(6,5)$ for the candidate image, whose vector is denoted by $\vec{c}_{6,5}$. We start by finding in the query image the best match for $\vec{c}_{6,5}$, which happens to be at location $(7,4)$ and is denoted by $\vec{q}_{7,4}$. We call these two locations \textit{anchor points}. 

Our spatial matching approach is  based on evaluating  location consistency of matching vectors in both images with respect to the anchor points. In the figure, if the two example anchors represent the same reality, then we would expect that the best match for the vector on the left-hand side  of the candidate's anchor, $\vec{c}_{6,4}$,  would be   $\vec{q}_{7,3}$ in the query image. Similarly, the best match for the vector on the right-hand side of the anchor, $\vec{c}_{6,6}$,  would be   $\vec{q}_{7,5}$. We continue  checking this consistency for the rest of the row in which the anchor is located. As exemplified  in Figure\,\ref{fig_spatial_matching} and  in order to explore the vertical direction, we also check for consistency among the anchor's column. We carry out the above procedure not only for position $(6,5)$, but for every single position in the candidate image. For each candidate, we accumulate a score counting the number of successfully matched locations over the rows and columns of all  positions and store them in a histogram  of candidates.  
Finally, we choose from the histogram the bin with the greatest accumulated score as the place match for the  query.

\begin{algorithm}[t!]
\caption{Spatial matching stage}
\begin{algorithmic}[1]
\small
\State $candidates \gets$ Get candidate list from Algorithm 1
\State $\{\vec{q}_{i,j}\} \gets$ Get query image vectors from $conv4\_2$ 
\State $cand\_hist \gets$ Initialize histogram of candidate images
\ForEach {candidate in $candidates$} 
	\State $\{\vec{c}_{i,j}\} \gets$ Get candidate vectors from SMDB
	\ForAll{$i,j\in\{0\dots12\}$}
		\State $\vec{q}_{k,l} \gets$ Find best match of $\vec{c}_{i,j}$ in query image
		\State Set $(i,j)$ and $(k,l)$ as anchor points
		\For {$n=-j$ to $(12-j)$} (scan current row)
			\If {$\vec{q}_{k,l+n}$ is best match for $\vec{c}_{i,j+n}$}
				\State Increase candidate bin in  $cand\_hist$   
			\EndIf
		\EndFor 
			\For {$m=-i$ to $(12-i)$} (scan current column)
			\If {$\vec{q}_{k+m,l}$ is best match for $\vec{c}_{i+m,j}$}
				\State Increase candidate bin in  $cand\_hist$  
			\EndIf
		\EndFor 
	\EndFor
\State Select from $cand\_hist$	 the candidate with the highest bin 	
\end{algorithmic}
\label{alg2}
\end{algorithm}
\section{EXPERIMENTS}\label{results}
\subsection{State of the art in VPR}
We compared our results with published performance of various VPR approaches. In particular, we considered the recent methods developed in \cite{chen2017only} (Region-BoW) and \cite{khaliq2018holistic} (Region-VLAD) as well as the approaches against which they compared their work (see \cite{chen2017only} and \cite{khaliq2018holistic} for an explanation): \textit{FAB-MAP}, 
  \textit{SeqSLAM},
 \textit{AlexNet},
 \textit{SUMPOOL},
 \textit{MAXPOOL} 
 and \textit{CROSSPOOL}.
  
\subsection{Datasets}\label{datasets}
\begin{figure}[htbp]
\vspace*{-3mm}
\hspace*{-3mm}
\includegraphics[height=10.5cm, width=9.0cm]{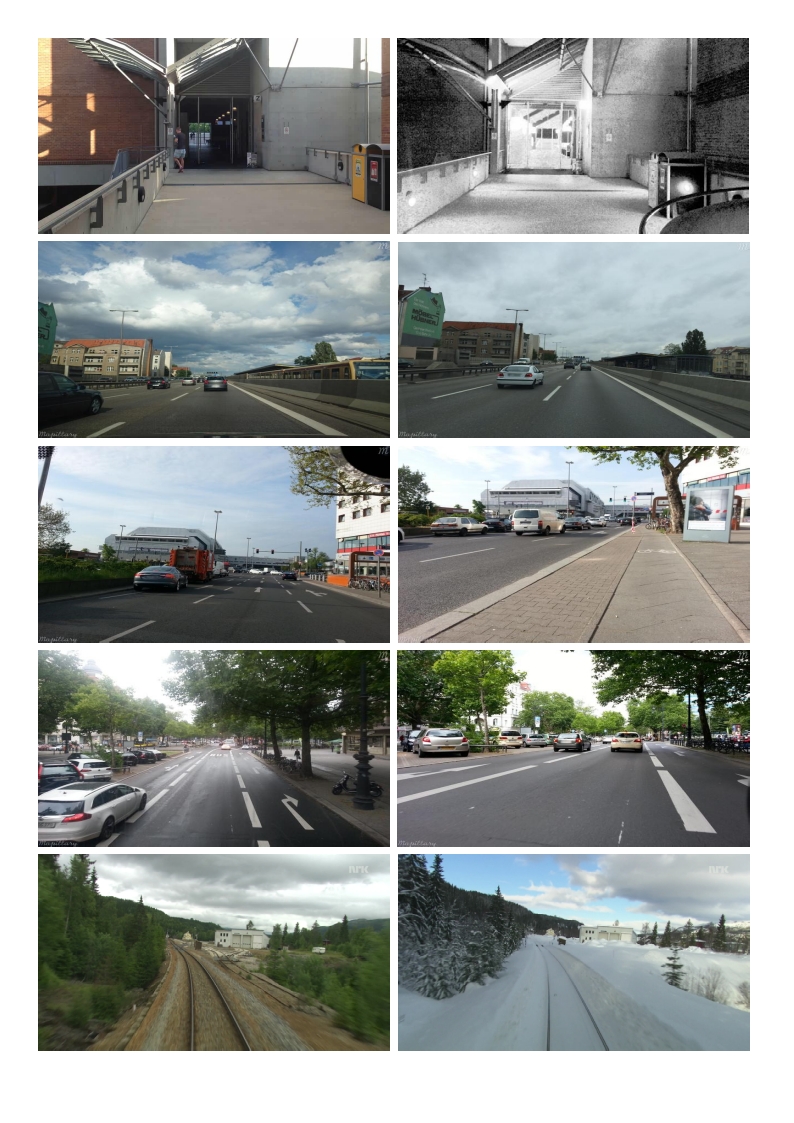}
\vspace*{-9mm}
\caption{Same place examples of reference (left) and query (right) images for the datasets considered in this work. Rows are, in descending order, \textit{Gardens Point}, \textit{Berlin A100}, \textit{Halenseestrasse}, \textit{Kudamm} and \textit{Nordland} datasets.} 
\vspace*{-5mm}
\label{fig_images}
\end{figure}

The datasets employed throughout this work are the same or similar to those used by\,\cite{khaliq2018holistic} and \cite{chen2017only}, in an effort to make meaningful comparisons of our results with those of the recent literature. They are summarized in Table~\ref{tbl_datasets}.
\begin{table}[h]
\vspace{-1mm}
\caption{Summary of the datasets used throughout this work}
\vspace{-3mm}
\label{tbl_datasets}
\begin{center}
\begin{tabular}{|l|l|l|}
\hline
\bf{Dataset} & \bf{Ref. images} &\bf{ Query  images}   \\
\hline
Gardens Point & \,\,\,200 (day-left)& \,\,\,200 (night-right)\\
\hline
Berlin A100 & \,\,\,\,\,\,85 (car) & \,\,\,\,\,\,81 (car)\\
\hline
Berlin Halenseestrasse & \,\,\,\,\,\,67 (car) & \,\,\,157 (bicycle) \\
\hline
Berlin Kudamm & \,\,\,201 (bus) & \,\,\,221 (bicycle)\\
\hline
Synthesized Nordland & 1415 (summer) & 1415 (winter) \\
\hline
\end{tabular}
\end{center}
\vspace*{-3mm}
\end{table}
These datasets have been well described elsewhere and represent a wide range of environments, viewpoint variations and condition changes. 
We refer the reader for instance to \cite{chen2017only, khaliq2018holistic, sunderhauf2015performance, sunderhauf2015place} for a more thorough description. The datasets themselves were downloaded as referenced in \cite{khaliq2018holistic}.
Figure~\ref{fig_images} shows some examples of the reference and query images for the datasets considered.

\subsection{Results and discussion}
\begin{figure}[t!]
\vspace*{-4mm}
\centering
\includegraphics[scale=0.43]{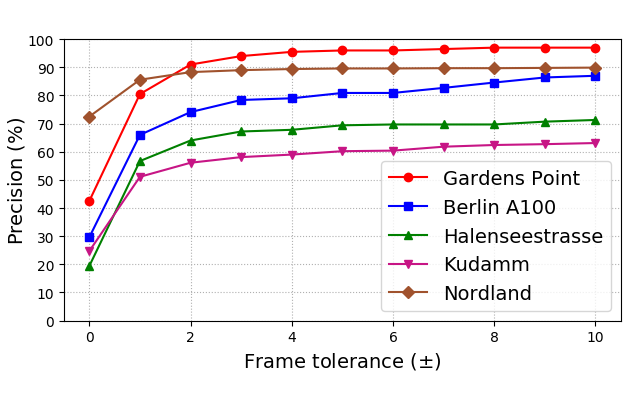}
\vspace*{-5mm}
\caption{Recognition performance as a function of the ground truth frame tolerance.} 
\label{fig_prec_vs_frame_tol}
\end{figure}

Before presenting our recognition results, we justify the choice of ground truth tolerance  used in our experiments.  Results for the analysis mentioned in Section~\ref{ground_truth} are presented in Figure~\ref{fig_prec_vs_frame_tol}. They show the recognition precision at 100\,\% recall for $d\!=\!100$ and $N\!=\!50$ when considering several frame tolerances in the assignation of true positives. Zero tolerance signifies that only those query images whose guessed location is exactly the ground truth  are considered true positives. 

As can be seen in the figure, relaxing the tolerance quickly translates into an increase in precision. This is because there may be several query images whose guessed place is not exactly the ground truth but fairly close to it and therefore they can be considered as true positives. We also see in the figure that above certain tolerances the performance gain rate becomes both small and constant. This is consistent with the random inclusion of images as true positives due exclusively to larger tolerances but otherwise uncorrelated with the current place. We consider this drop in performance gain as an intrinsic  indicator of the spatial limits of a given place and consequently we    set the frame tolerance to $\pm\,2$ frames. In all datasets, this value  meets the distance threshold of 25\,m (Section \,\ref{ground_truth}) although for  the sake of completeness, we will also provide some results for tolerances of $\pm\,1$ and $\pm\,3$ frames.

\begin{figure}[t!]
\vspace*{-5mm}
\centering
\includegraphics[scale=0.43]{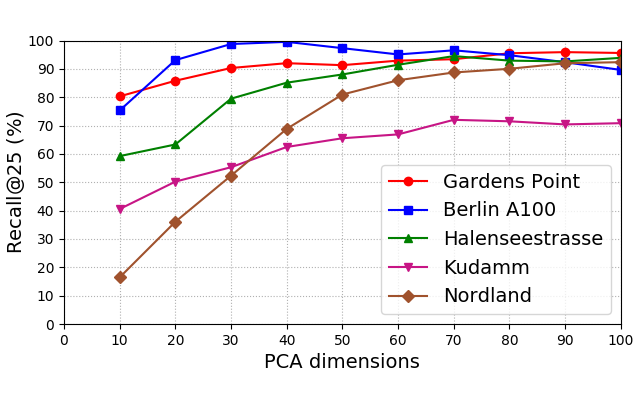}
\vspace*{-5mm}
\caption{Effect of the number of dimensions on image filtering performance (recall@25) for layer  \texttt{block5\_conv2}.} 
\label{fig_pca_comp}
\end{figure}

We present in Figure~\ref{fig_pca_comp} our analysis on the image filtering performance  as a function of the number of principal components. The results are given for a recall@25. We clearly see that for all datasets there is no significant gain in recall above around 80-90 components. This allowed us to reduce the CNN feature vectors to 100 dimensions without apparent lost in performance. We also made this assumption for the spatial matching stage (a detailed analysis like the one in the figure will be provided in future work).
\begin{figure}[t!]
\vspace*{-3mm}
\centering
\includegraphics[scale=0.43]{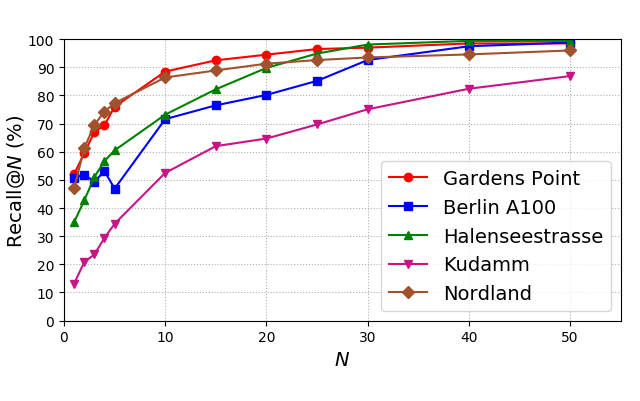}
\vspace*{-5mm}
\caption{Recall as a function of the number of candidates $N$ in the image filtering stage.} 
\vspace*{-5mm}
\label{fig_recall_vs_N}  
\end{figure}  

With the number of dimensions fixed to $d\!\!=\!\!100$, we investigated the influence of the candidate list size $N$ on the image filtering recall performance, which is summarized in Figure~\ref{fig_recall_vs_N}. For $N\!\!=\!\!1$, the recall can be interpreted as the place recognition performance of layer \textit{conv\,5-2}, as only the best candidate is taken into account.
The performance of this layer is comparable for instance to CNN methods that use the whole image\,\cite{sunderhauf2015performance, sunderhauf2015place}.

As the number of candidates becomes larger,  we see that the recall increases rapidly,  although the growth progressively slows down for larger values of $N$. With the exception of the \textit{Kudamm} dataset, which is known to be particularly challenging,  the recall is above 95\% for all datasets at $N\!=50$, justifying the use of this value in our experiments.
\begin{figure}[t!]
\vspace*{-0mm}
\centering
	\includegraphics[scale=0.234]{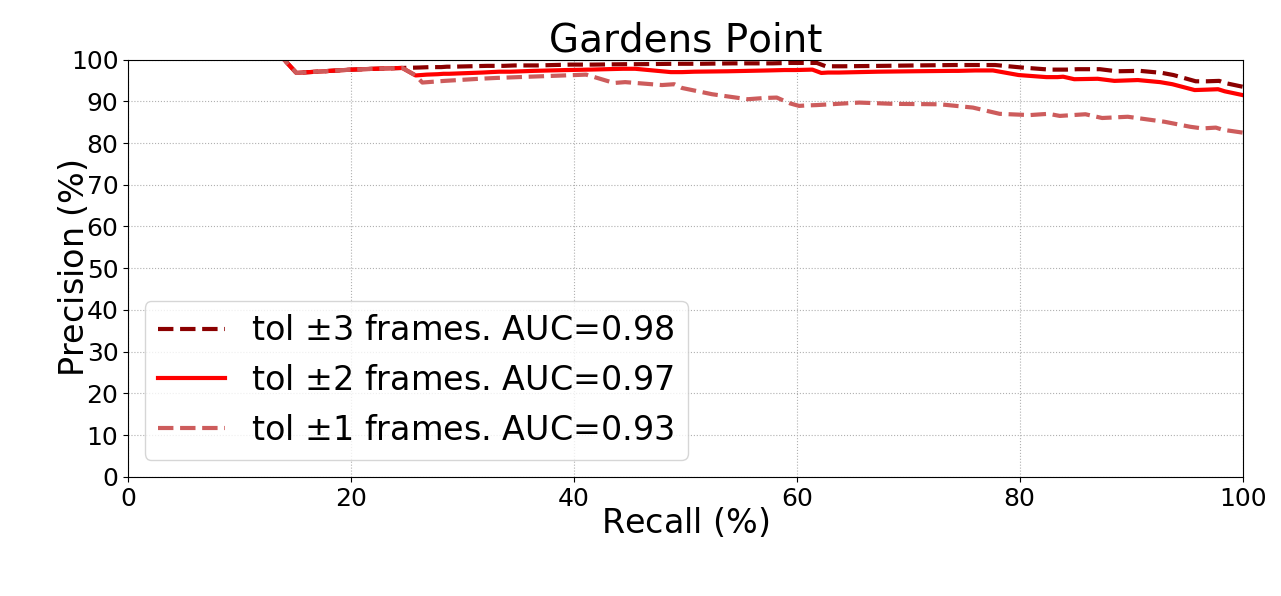} \\
	\includegraphics[scale=0.234]{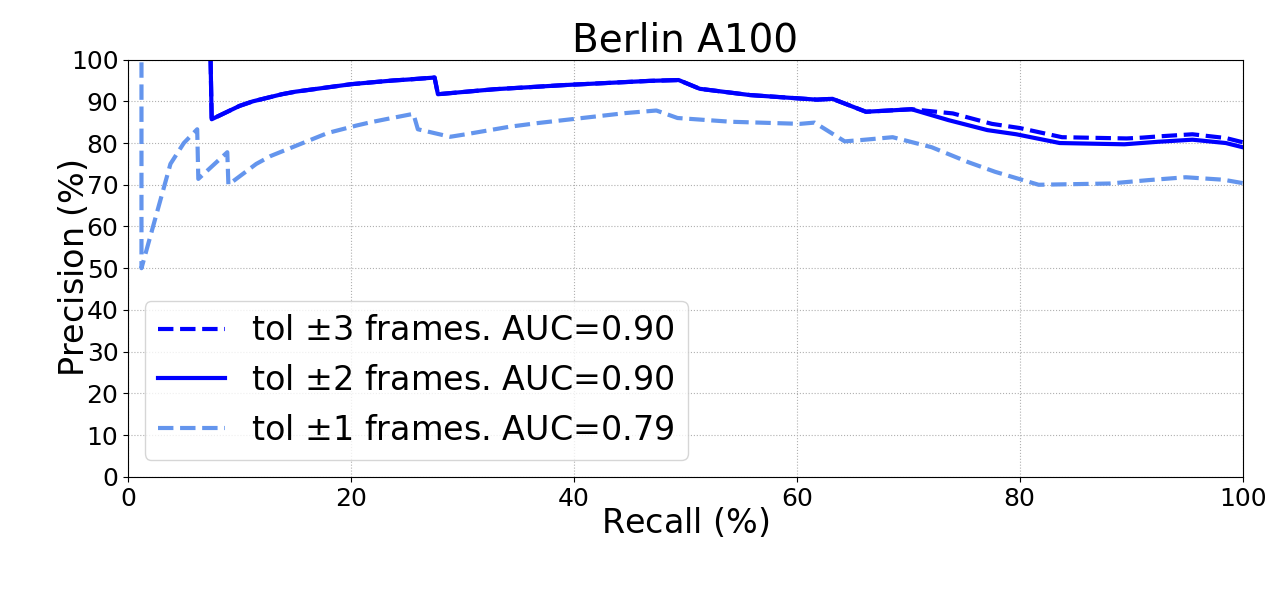}\\
	\includegraphics[scale=0.234]{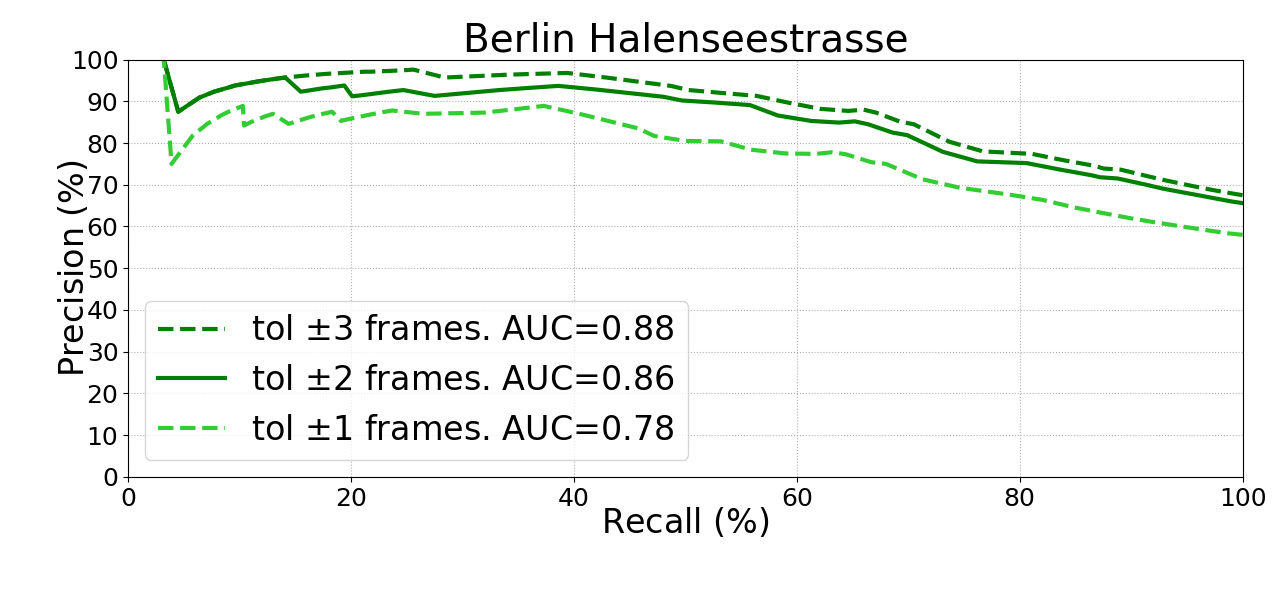}\\
	\includegraphics[scale=0.234]{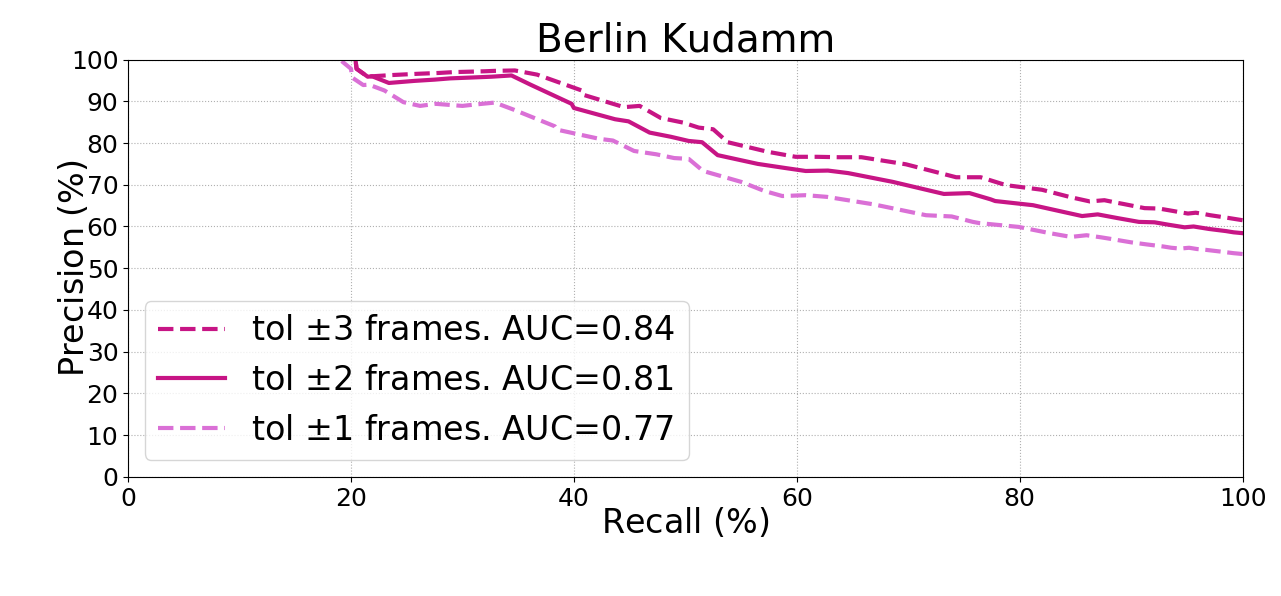} \\
	\includegraphics[scale=0.234]{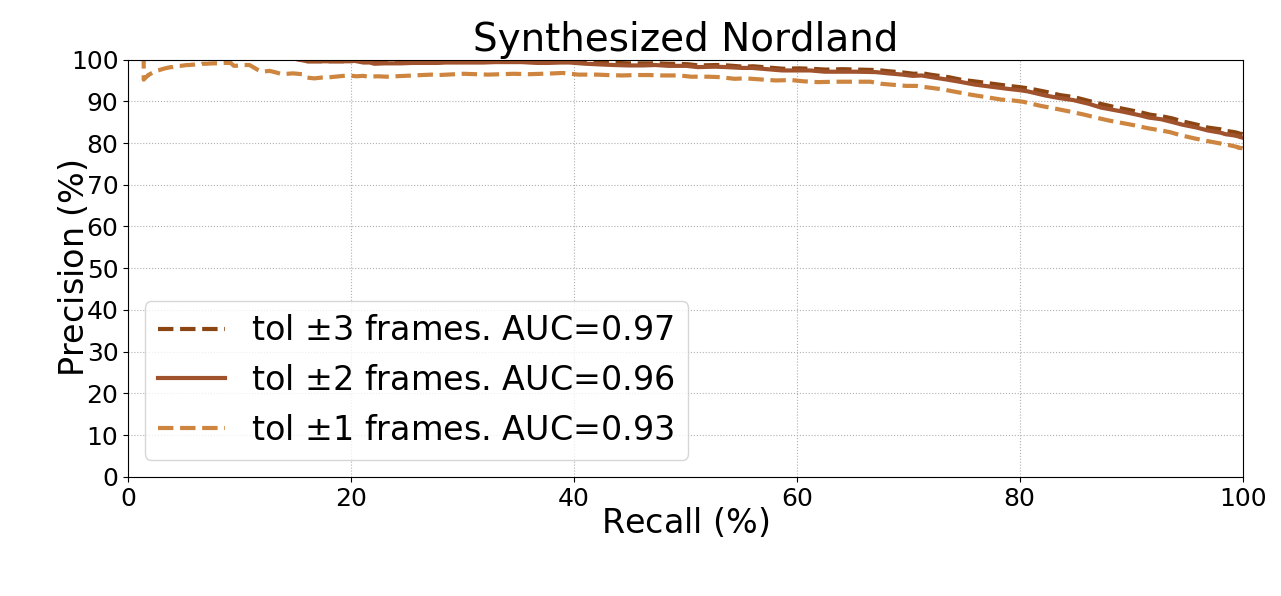} 
\caption{Precision-recall curves for 3 ground truth frame tolerances.} 
\vspace*{-5mm}
\label{fig_prec-recall_curves}
\end{figure}  
\begin{figure}[t!]
\vspace*{-0mm}
\centering
	\includegraphics[scale=0.232]{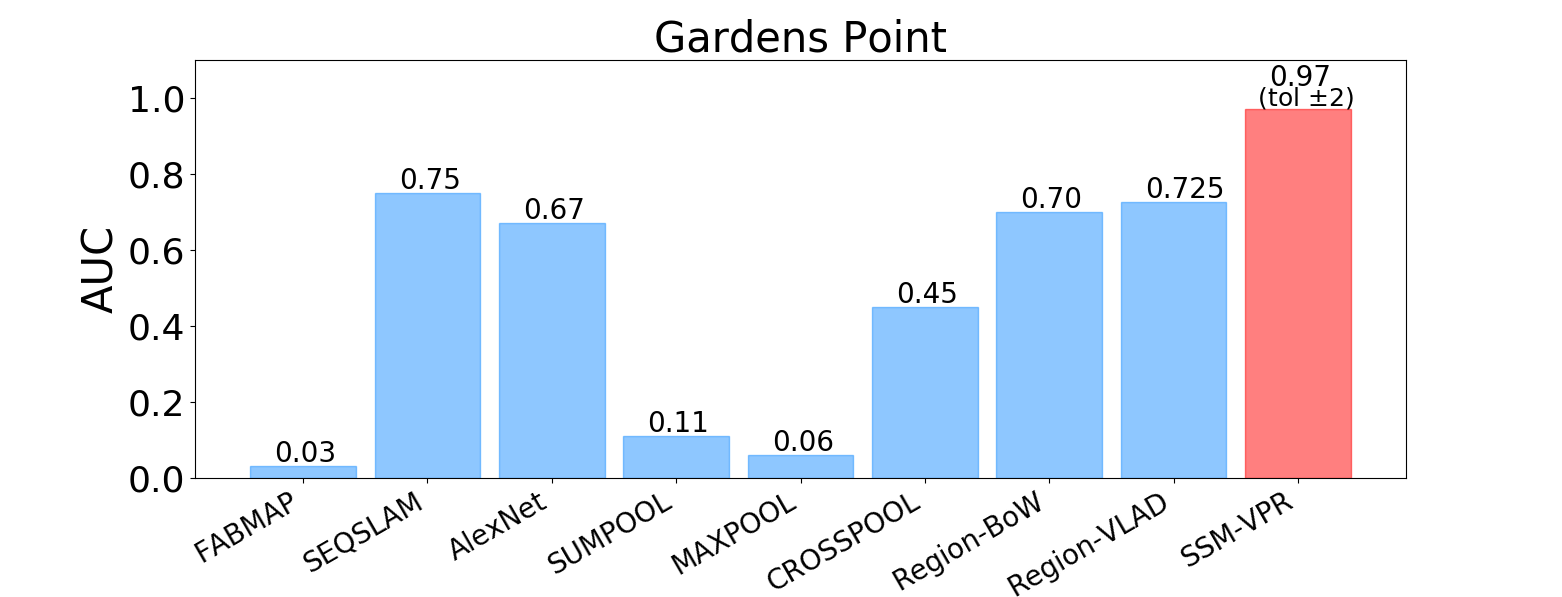} \\
	\includegraphics[scale=0.232]{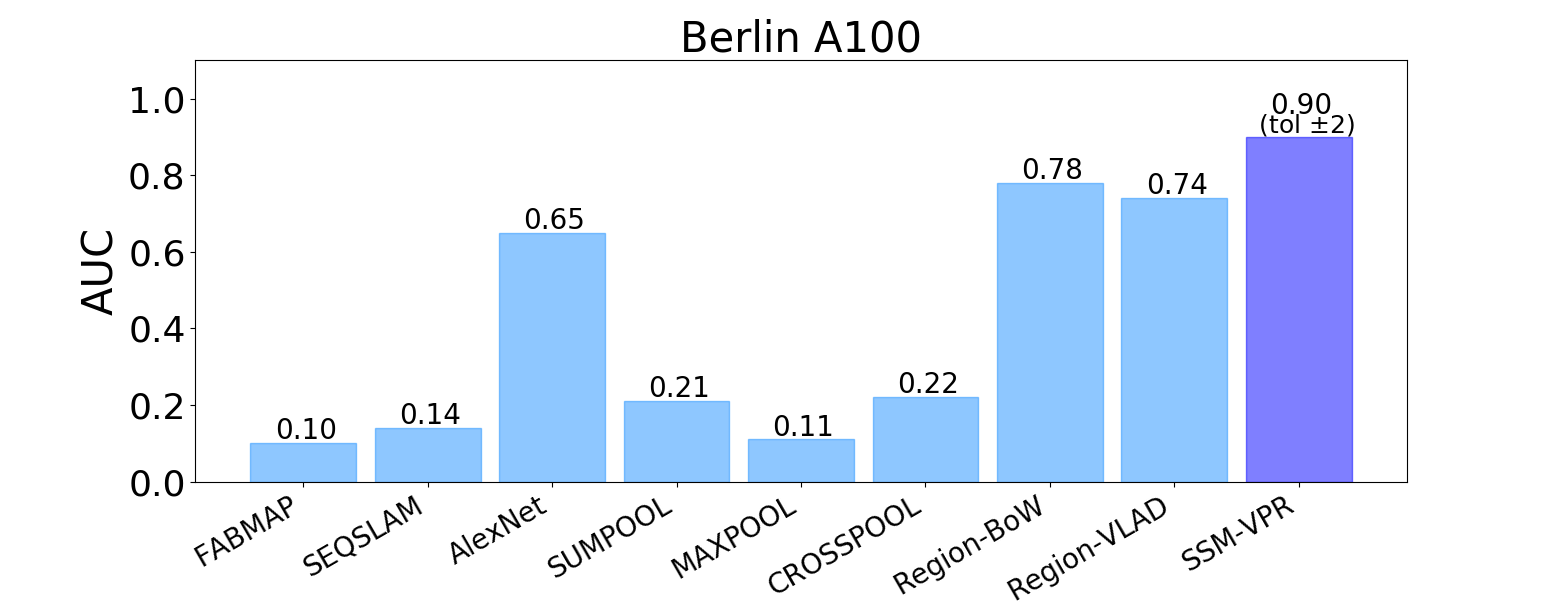}\\
	\includegraphics[scale=0.232]{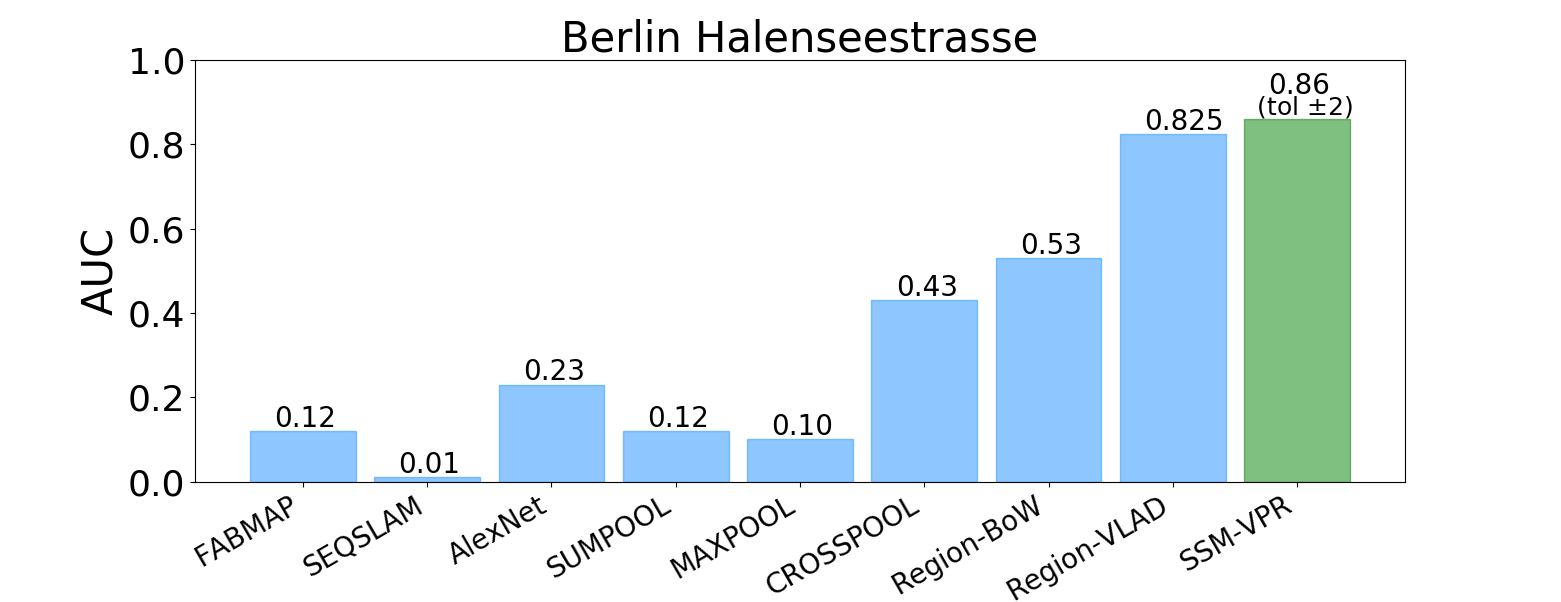}\\
	\includegraphics[scale=0.232]{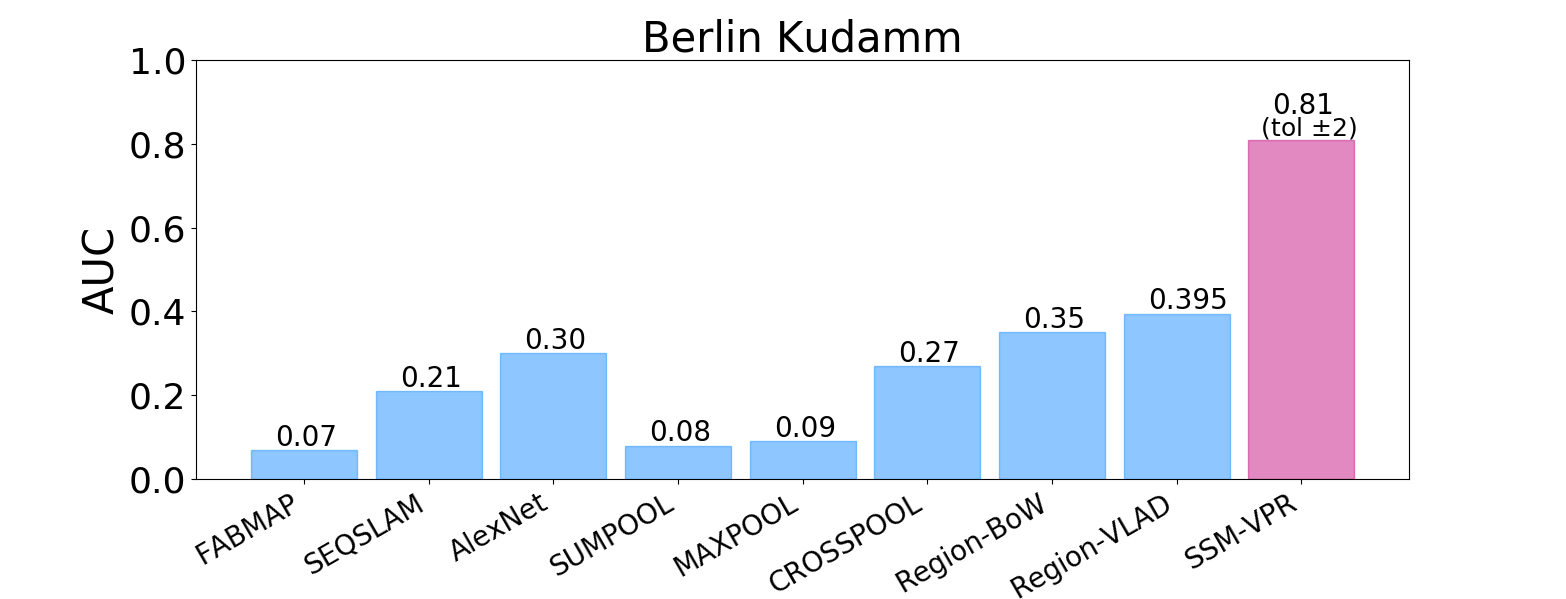} \\
	\includegraphics[scale=0.232]{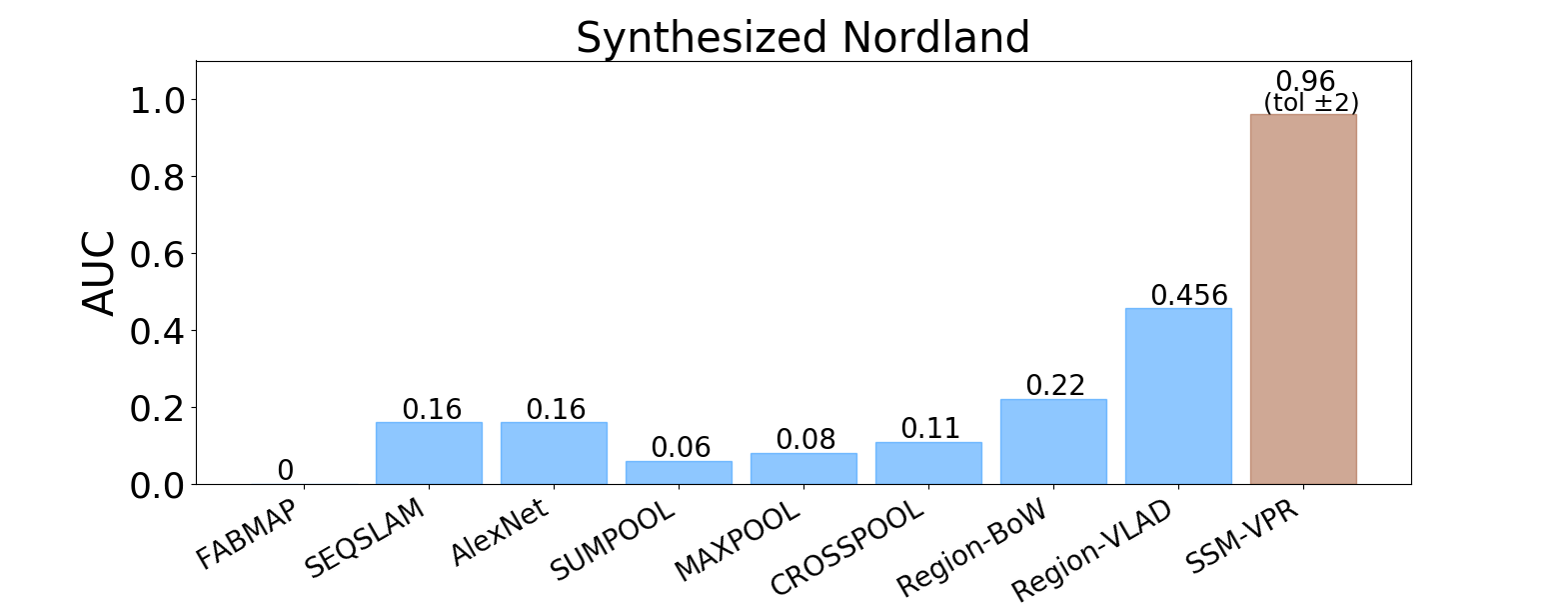} 
\caption{Area Under the Curve (AUC) of the precision-recall curves.} 
\vspace*{-4mm}
\label{fig_auc_curves}
\end{figure} 

We assessed the recognition performance of our two-stage system on the five datasets by evaluating precision-recall curves on them. The resulting curves are shown in Figure~\ref{fig_prec-recall_curves}, where three ground truth frame tolerances are considered for completeness. 
The curves show that taking a tolerance of $\pm1$ is suboptimal, as a significant boost in performance is achieved in all cases by 
increasing it to $\pm2$. On the other hand, relaxing this constraint to $\pm3$ frames only improves performance by a small amount, a fact that supports our choice of $\pm2$ as the standard frame tolerance for the identification of a place. 

As can be seen in Figure~\ref{fig_auc_curves}, we compared our recognition results with recent published work\,\cite{khaliq2018holistic, chen2017only} by using the Area Under the Curve (AUC) of the precision-recall curves.  On the five benchmark datasets considered, we consistently outperformed  visual place recognition results. For \textit{Gardens Point} and \textit{Synthesized Nordland}, which represent very strong condition and viewpoint changes for the former and strong condition changes for the latter, we are not too far from  perfect performance,  with an AUC\,=\,0.97. To the best of our knowledge, this level of recognition is unprecedented on these datasets. On the \textit{Synthesized Norland}, the improvement with respect to the state-of-the-art is twofold. It is also remarkable the excellent performance obtained on the \textit{Kudamm} dataset, which is considered in the literature as very challenging. Our results show again a performance that is  two times better than the best published results on this dataset. Notice that even if a frame tolerance of $\pm1$ is considered, we are still on par with  published results for the Berlin A100 and Halenseestrasse datatests and  well above for the rest, making clear the superiority of our method.

\subsection{Runtime and memory considerations}
At this stage of our work, we have not excessively focused on computational complexity, since our goal was on recognition performance. The current system can find a place in an average of 1.6 seconds. There exist however several opportunities for reducing this time considerably. The most clear one is to reduce the number of candidate images retrieved in the image filtering stage. Currently, we are using 50 images. A reduction to 15 candidates brings down recognition time to 750\,ms and 400\,ms for a candidate list of 5 images. With an improvement in our image filtering stage, this could lead to a very competitive recognition system. Regarding the image filtering and spatial matching databases, each image takes approximately 25\,kB and 150\,kB of disk space, respectively.

\section{CONCLUSIONS AND FUTURE WORK}\label{conclusions}
In this work, we have shown that CNN-based image database filtering  combined with exhaustive spatial matching of filtered candidates is a successful two-stage approach to the problem of visual place recognition. Following the success of CNNs in several computer vision tasks, we used both their power creating robust representations of images as well as their spatial properties to tackle recognition. We showed that using this combination significantly increases the published recognition performance in commonly used benchmark datasets, some of them highly challenging. It is precisely on these that we achieved the most striking results, with more than a twofold increase in performance with respect to the state-of-the-art.  

Our immediate efforts will be focusing  on improving the image filtering stage of our system.  In particular, we will look at end-to-end trainable architectures such as NetVLAD as the most straightforward  approach to both increase the recognition performance and reduce the complexity of the system.

Furthermore, it is our intention to explore alternative spatial matching algorithms, experiment with different image resolutions and introduce knowledge about previously recognized frames to leverage the frame correlation inherent to sequences of images when moving through places. 

Finally, we will try to expand the number of datasets in order to include more indoor as well as countryside scenarios and we will start to integrate our system in real robotic applications, with a view on FPGA real-time implementations.

\section*{ACKNOWLEDGMENT}
This work has been supported by the European Union's Horizon 2020
research and innovation programme under grant agreement No 688117, by
the Technology Agency of the Czech Republic under the project
no.~TE01020197 \enquote{Centre for Applied Cybernetics}, and
by the European Regional Development Fund under the project Robotics for
Industry 4.0 (reg. no. CZ.02.1.01/0.0/0.0/15 003/0000470)

\printbibliography

\end{document}